\documentclass[runningheads]{llncs}
\usepackage[T1]{fontenc}

\usepackage{graphicx}
\usepackage{hyperref}
\usepackage{multirow}
\usepackage{amssymb}
\usepackage{amsmath}
\usepackage{bm}
\usepackage{pifont}
\usepackage{pgfplots}
\pgfplotsset{compat=1.17}
\usepackage{fontspec} 
\usepackage{adjustbox} 

\begin{document}
\title{EnsemJudge: Enhancing Reliability in Chinese LLM-Generated Text Detection through Diverse Model Ensembles}
\titlerunning{EnsemJudge}
%
\author{Zhuoshang Wang\inst{1,2} \and
Yubing Ren\inst{1,2}\thanks{Corresponding author.} \and
Guoyu Zhao\inst{1,2} \and \\
Xiaowei Zhu\inst{1,2} \and
Hao Li\inst{1,2} \and
Yanan Cao\inst{1,2}}
\authorrunning{Z. Wang et al.}
\institute{Institute of Information Engineering, Chinese Academy of Sciences, Beijing, China \and
School of Cyber Security, University of Chinese Academy of Sciences, Beijing, China\\
\email{\{wangzhuoshang,renyubing\}@iie.ac.cn}}
\maketitle              
\begin{abstract}
Large Language Models (LLMs) are widely applied across various domains due to their powerful text generation capabilities. While LLM-generated texts often resemble human-written ones, their misuse can lead to significant societal risks. Detecting such texts is an essential technique for mitigating LLM misuse, and many detection methods have shown promising results across different datasets. However, real-world scenarios often involve out-of-domain inputs or adversarial samples, which can affect the performance of detection methods to varying degrees. Furthermore, most existing research has focused on English texts, with limited work addressing Chinese text detection. In this study, we propose EnsemJudge, a robust framework for detecting Chinese LLM-generated text by incorporating tailored strategies and ensemble voting mechanisms. We trained and evaluated our system on a carefully constructed Chinese dataset provided by NLPCC2025 Shared Task 1. Our approach outperformed all baseline methods and achieved first place in the task, demonstrating its effectiveness and reliability in Chinese LLM-generated text detection. Our code is available at \url{https://github.com/johnsonwangzs/MGT-Mini}.

\end{abstract}
\section{Introduction}
Large Language Models (LLMs) have become increasingly integrated into daily life due to their impressive natural language generation capabilities. As LLMs continue to evolve and become more accessible, the cost of generating text has significantly decreased, leading to growing concerns over the misuse of LLM-generated content. For instance, students may use LLMs to write academic papers on their behalf, malicious actors might produce and disseminate harmful fake news, and hallucinated content from LLMs could spread misinformation, negatively impacting online communities. These issues pose serious challenges to the responsible development and deployment of LLMs.

As a result, detecting LLM-generated text has become an urgent and essential task. Early approaches were primarily rule-based, relying on the extraction of specific statistical features from the text for classification~\cite{gehrmann-etal-2019-gltr,hashimoto-etal-2019-unifying,ippolito-etal-2020-automatic}. However, with the advancement of LLMs, LLM-generated text has become increasingly similar to human-written content, rendering such statistical methods progressively less effective. In recent years, training-based methods~\cite{solaiman2019release,NEURIPS2023_30e15e59,NEURIPS2024_a117a3cd,NEURIPS2024_1d35af80,NEURIPS2024_bc808cf2,song2025deep}—such as fine-tuning pretrained language models—and  training-free approaches~\cite{pmlr-v202-mitchell23a,DBLP:conf/iclr/Yang0WPWC24,bao2024fastdetectgpt,hans2024spotting,DBLP:journals/corr/abs-2410-06072} based on generation probability have emerged. Nevertheless, these methods often suffer from notable limitations: rule-based approaches are labor-intensive and rely heavily on handcrafted rules; training-based methods tend to struggle with generalization to out-of-domain data; and training-free methods are sensitive to factors such as text length, making them less reliable for real-world deployment. Moreover, the majority of existing work has been conducted on English datasets, and the effectiveness of these methods on other languages, such as Chinese, remains underexplored.

In this paper, we propose an ensemble-based framework for LLM-generated text detection, aiming to optimize overall detection performance across diverse scenarios. Specifically, we treat each existing detection method as an independent base model and evaluate their performance differences when handling various types of text. We dynamically assign ensemble strategies to different inputs, each strategy applies differentiated voting weights to the individual models, thereby enabling the system to achieve optimal performance in aggregate. This framework effectively leverages the strengths of different models on different text types, ensuring robust and reliable performance across many real-world conditions.

Based on this framework, we develop a complete LLM-generated text detection system. We participate in the NLPCC2025 Shared Task 1, where we train and fine-tune our models using the official training set. Evaluation results show that our system achieves the highest macro F1 score compared to baseline models on the test set, which includes normal and adversarial texts. It is important to note that the test set differs significantly from the training set in terms of data distribution, and its construction methodology is entirely unknown to participants before the release of evaluation results. As a result, system optimization relies solely on anticipating potential adversarial strategies.

Our main contributions are as follows:
\begin{enumerate}
\item We evaluate the performance of several mainstream LLM-generated text detection methods on Chinese text data and assess their robustness against a variety of adversarial scenarios that may arise in real-world applications.
\item We propose an ensemble-based detection framework, EnsemJudge, which effectively integrates the strengths of multiple models by applying tailored strategies and ensemble voting rules to different types of text.
\item Based on this framework, we implement an extensible LLM-generated text detection system which achieves first place on the NLPCC2025 Shared Task 1, demonstrating its strong performance and reliability.
\end{enumerate}

\section{Related Work}
Detecting LLM-generated texts has attracted increasing attention due to its critical role in enhancing transparency and preventing misuse. Existing detection methods primarily include training-based methods and training-free methods.

\noindent\textbf{Training-based methods} typically utilize supervised classifiers to differentiate human-written from LLM-generated texts, such as early RoBERTa classifiers \cite{solaiman2019release}. Recent advancements like RADAR \cite{NEURIPS2023_30e15e59}, DeTeCtive \cite{NEURIPS2024_a117a3cd}, DPIC \cite{NEURIPS2024_1d35af80}, and Biscope \cite{NEURIPS2024_bc808cf2} leverage adversarial training, contrastive learning, prompt reconstruction, and statistical features, respectively. However, these methods face significant limitations, particularly poor generalization to out-of-distribution (OOD) scenarios due to feature overfitting \cite{chakraborty2023possibilitiesaigeneratedtextdetection}, prompting a shift toward training-free solutions.

\noindent\textbf{Training-free methods} exploit statistical and probabilistic text characteristics, bypassing model training. Techniques such as LogRank \cite{gehrmann-etal-2019-gltr}, Likelihood \cite{hashimoto-etal-2019-unifying}, and Entropy \cite{ippolito-etal-2020-automatic} analyze uncertainty metrics. DetectGPT \cite{pmlr-v202-mitchell23a} introduced contrastive perturbation paradigms, later enhanced by methods like Fast-DetectGPT \cite{bao2024fastdetectgpt} for improved efficiency, Binoculars \cite{hans2024spotting} using cross-model perplexity, and Lastde++ \cite{DBLP:journals/corr/abs-2410-06072} via Diversity Entropy. Despite advancements, computational efficiency and real-time detection remain key challenges.

\section{Method}
In this section, we present EnsemJudge, an ensemble-based framework for LLM-generated text detection. We begin by outlining the motivation for adopting an ensemble approach, followed by a detailed description of the framework.
\subsection{Observation and Motivation}
Although numerous LLM-generated text detection methods have been proposed in recent years~\cite{wu2025survey}, most of them are designed and evaluated on English datasets, with limited evidence of their effectiveness on Chinese text. Therefore, we first extract features or train models based on the training set and evaluate the performance of various detection methods on the development set.\footnote{According to the guidelines of NLPCC2025 Shared Task 1, all data involved in the task are in Chinese, and participants are restricted to using only the officially provided training and development sets for system development; the use of external data is not permitted~\cite{nlpcc2025task1}.}

\subsubsection{Data Augmentation.}
Based on existing data, we construct adversarial examples targeting potential evasion strategies, in order to simulate challenges that may arise in real-world LLM-generated text detection scenarios. We design two main types of adversarial methods. The first is back-translation, where an encoder-decoder model is used to translate the text generated by a LLM from the source language to a target language and then back to the source language. Prior studies have shown that detection accuracy drops significantly under paraphrasing attacks; thus, we use this method to simulate such attacks. The second method is text excerpting. We observe that many commercial detection tools reject inputs shorter than a minimum threshold, indicating that short-text detection remains a critical challenge.

\subsubsection{Detection Method Evaluation.}
Given that our task targets LLM-generated text detection in practical application settings, we select a set of representative detection approaches for evaluation. Details of their implementation can be found in Section 4.1.

\paragraph{Rule-based Methods.} 
Rule-based methods are primarily used to detect common character patterns in text. In LLM-generated text, certain special tokens (e.g., \textbackslash n\textbackslash n) and frequently used phrases are more likely to appear. In contrast, human-written text tends to contain more clauses separated by commas and may exhibit more informal writing patterns, such as repeated punctuation marks. Although these approaches are mostly “ad-hoc”, they can be highly effective when relevant patterns are present, often playing a decisive role in detection.

\paragraph{Training-free Methods.}
We adopt three training-free detection methods, including two state-of-the-art approaches—Binoculars~\cite{hans2024spotting} and Fast-DetectGPT~\cite{bao2024fastdetectgpt}—as well as a self-designed method named CommonToken. For Fast-DetectGPT, we experiment with different backbone models and both the normal and analytical modes. CommonToken operates by computing token frequency statistics from the positive and negative subsets of the training data. Given an input text, it simply compares the number of tokens originating from each subset and makes a prediction accordingly

\paragraph{Training-based Methods.}
We employ three types of training-based methods. Following the work of Guo et al.~\cite{guo2023closechatgpthumanexperts}, we fine-tune Chinese versions of RoBERTa and BERT~\cite{cui-etal-2020-revisiting} as classifiers. To evaluate the inherent capability of LLMs in detecting LLM-generated text, we also perform instruction tuning using LoRA on several Chinese LLMs. Additionally, building on the previously mentioned rule-based and training-free methods, we construct a hybrid feature model by concatenating the output representations from RoBERTa with the feature outputs of other detection methods. We then train a classification head on top of these combined features. We believe this integration helps leverage the strengths of both types of approaches.

\subsubsection{Observation and Intuition.}
Based on the evaluation results, we make several key observations. It is worth noting that the test set released by the task organizers is more comprehensive than the evaluation set we construct from the training data in the early stage of the competition. Furthermore, we observe that the performance of various methods remains highly consistent across our self-constructed evaluation set and the final test set. Due to space limitations, we present only the key observations here; for detailed results and analysis, please refer to Section 4.2.
\begin{enumerate}
\item Several methods that perform well on English datasets exhibit noticeable performance degradation on Chinese datasets.
\item The effectiveness of each method varies across different types of text; individual methods tend to specialize in handling specific categories of content.
\item Detection methods show differing biases or tendencies when making binary decisions (i.e., classifying text as LLM-generated or human-written).
\item Methods of the same category tend to exhibit similar performance patterns; however, their actual effectiveness heavily depends on implementation details.
\item Certain adversarial strategies, such as back-translation or excerpting short text segments, can easily degrade the effectiveness of many detection methods.
\end{enumerate}

Based on the above observations, we arrive at the following intuition: although an individual detection method may perform well on certain types of data, it is unlikely to maintain strong performance across all scenarios. Given the diversity of text encountered in real-world applications, the most effective way to ensure overall reliability of the detection system is to construct an ensemble model that integrates multiple methods through carefully designed decision strategies to leverage their respective strengths.

\subsection{EnsemJudge: A Reliable Detection Framework}
We begin by defining the notion of reliability in LLM-generated text detection, followed by a detailed introduction to the EnsemJudge framework.

\subsubsection{Reliability in LLM-generated text detection.} Given a detection system $S$ and a set of texts $\mathcal{D}$ generated by LLMs. The goal of $S$ is to correctly predict whether a given text $t \in \mathcal{D}$ is human-written or LLM-generated. However, in real-world scenarios, the texts that $S$ must detect are often not directly generated by the models but may have undergone adversarial modifications. We denote the set of all possible adversarial transformations as $\mathcal{F}$, and the detection target becomes the transformed set of texts $\mathcal{D}'$, where $\mathcal{D}' = \{ f(t) \mid t \in \mathcal{D}, f \in \mathcal{F} \}$.

We define the reliability of a detection system $S$ for a text set $\mathcal{D}$ and a set of adversarial transformations $\mathcal{F}$ as the expected performance of $S$ over all adversarially modified texts. Formally, let $\text{Acc}(S, t)$ be an indicator function that returns 1 if $S$ correctly classifies $t$, and 0 otherwise. Then, the reliability $\text{Rel}(S, \mathcal{D}, \mathcal{F})$ is defined as:
\begin{equation}
\text{Rel}(S, \mathcal{D}, \mathcal{F}) = \mathbb{E}_{t \in \mathcal{D}, f \in \mathcal{F}}[\text{Acc}(S, f(t))]
\end{equation}
In practice, since enumerating all $f \in \mathcal{F}$ is infeasible, we approximate this expectation using a representative subset $\mathcal{F}' \subset \mathcal{F}$, such as selected text paraphrasing and excerpting attacks.

\subsubsection{The EnsemJudge Framework.} 
\begin{figure}[htbp]
  \centering
  \includegraphics[width=1\textwidth, trim=0cm 0cm 4cm 5cm, clip]{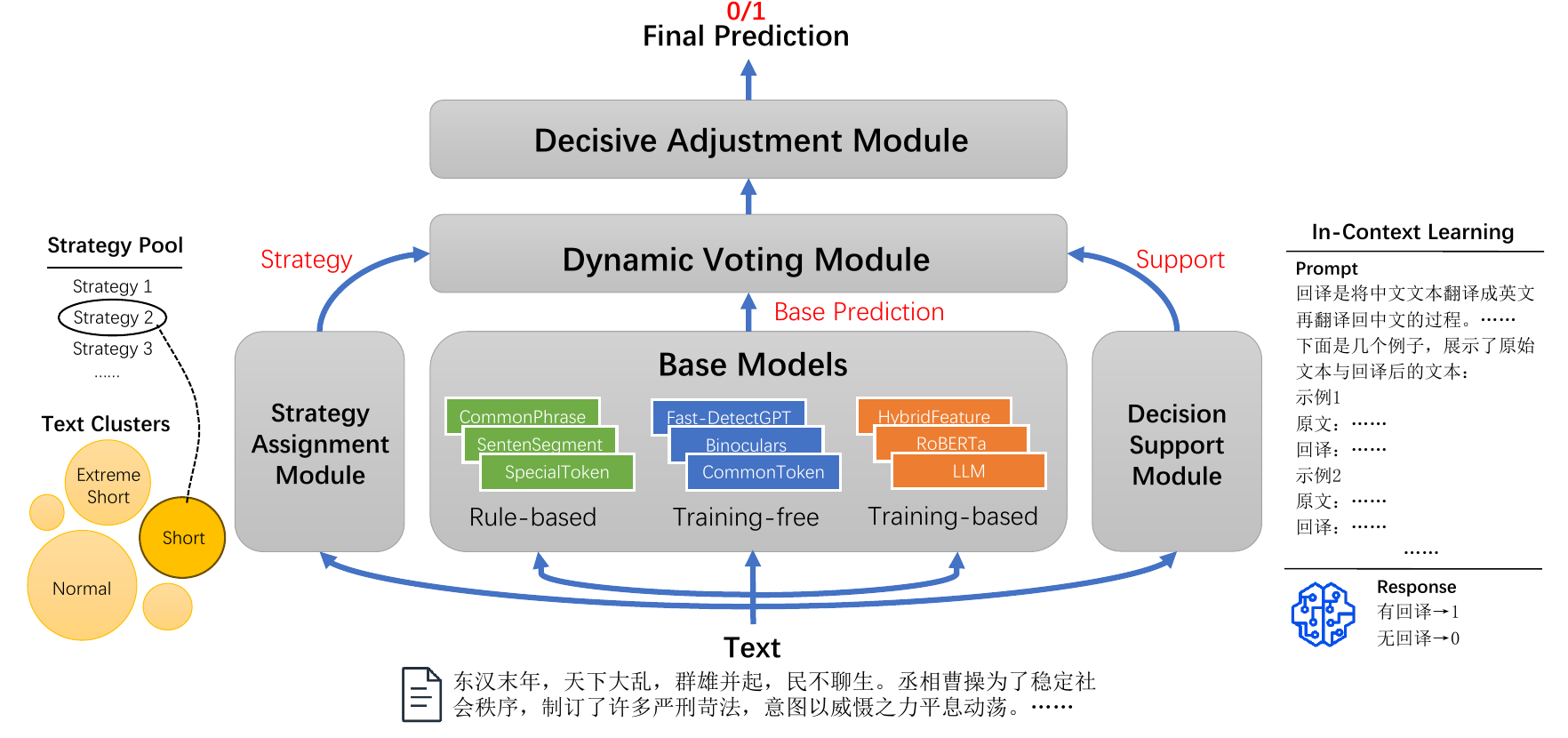}
  \caption{The EnsemJudge Detection Framework.}
  \label{fig:framework}
\end{figure}

We propose EnsemJudge, a reliable ensemble framework for LLM-generated text detection tailored to real-world scenarios. By dynamically assigning integration strategies to different texts and assigning differentiated voting weights to each detection method under each strategy, the framework aims to optimize overall system performance. EnsemJudge effectively leverages the strengths of various detection methods across different text types, thereby ensuring reliability in diverse settings. The overall architecture is illustrated in Figure~\ref{fig:framework}.

\vspace{-10pt}
\vspace{-20pt}

\paragraph{Base Models.}
We begin by selecting a set of existing LLM-generated text detection methods $\mathcal{M}$. Each method $M_i\in\mathcal{M}$, regardless of its complexity, is treated as an independent base detection model, from which we obtain predictions across the entire dataset. The framework adopts a modular design, allowing the integration of any detection method in principle, as long as its prediction output $y_{M}=M(\cdot)$ is binary (i.e., 0 or 1).

\paragraph{Strategy Assignment Module.} 
The strategy assignment module begins by analyzing the features of each input text $t \in \mathcal{D}$. Based on these features, it selects a subset of base models $\mathcal{M}^t = \{M_1^t, \cdots, M_n^t\}$ from the full set $\mathcal{M}$, and assigns predefined voting weights to the models within $\mathcal{M}^t$. Models not included in $\mathcal{M}^t$ do not contribute to the final decision.

The mapping from textual features to model subsets is determined in advance using a clustering-based approach. Specifically, we extract a set of lightweight features (e.g., text length, perplexity, etc.) for each text in the training set and perform clustering over these feature representations. For each resulting cluster, we evaluate the performance of all base models in $\mathcal{M}$, select an appropriate subset of models, and assign their voting weights to maximize detection performance—measured by metrics such as F1 score—on that cluster. The resulting voting configuration defines a strategy $r^{t} \in \mathcal{R}$, where $\mathcal{R}$ is the set of all learned strategies.

\paragraph{Decision Support Module.}
The decision support module is designed to handle samples that are difficult for the ensemble model to classify confidently. These samples typically yield voting scores near the decision threshold, resulting in higher uncertainty. This module relies on heuristic techniques to assist the ensemble decision-making process. For example, we incorporate large language models using in-context learning (ICL). Specifically, when the ensemble model exhibits uncertainty, we employ few-shot prompting to query the LLM and obtain a response $d^{t}$, which is then used to support or refine the prediction.

\paragraph{Dynamic Voting Module.}
We define the dynamic voting function as a weighted decision rule that combines base model predictions with optional adjustments from the decision support module. Specifically, for each input text $t \in \mathcal{D}$, let the voting strategy $r^t$ from the strategy assignment module define a weight vector $\boldsymbol{w}^t = \{w_{M_1}^t, \dots, w_{M_n}^t\}$, where each weight corresponds to a base model $M_i \in \mathcal{M}^t$. The initial ensemble prediction score is computed as:
\begin{equation}
s^t = \sum_{i=1}^{n} w_{M_i}^t \cdot v_{M_i}^t
\end{equation}
where $v_{M_i}^t \in \{-1, 1\}$ corresponds to the prediction of model $M_i$ for text $t$. The final prediction $y^t$ is determined by:
\begin{equation}
y^t =
\begin{cases}
1, & \text{if } s^t + \lambda \cdot d^t \geq \tau \\
0, & \text{otherwise}
\end{cases}
\end{equation}
Here, $d^t \in [-1, 1]$ is the support signal from the decision support module, $\lambda$ is a tunable parameter controlling the influence of the support module, and $\tau$ is the decision threshold.

\paragraph{Decisive Adjustment Module.}
The decisive module serves to make final adjustments to the voting results. In certain cases, LLM-generated or human-written texts exhibit highly distinctive features that can directly support classification. For example, double newline characters are typically only found in LLM-generated texts. However, this module also carries certain risks, as adversarial strategies can be designed to inject such representative features, potentially leading to misclassification. In our experiments, we observed that when the preceding ensemble voting component is sufficiently well-designed, this module can be safely omitted.

\section{Experiments}
In this section, we evaluate EnsemJudge for Chinese LLM-generated text detection based on the resources provided by NLPCC2025 Shared Task 1~\cite{nlpcc2025task1}. We first present the details of the dataset, evaluation metrics, and system implementation, followed by a report of the evaluation results.

\subsection{Experimental Settings}
\subsubsection{Datasets.}
The dataset used in this study is officially provided by the task organizers. It is divided into training, development, and test sets. Table~\ref{dataset} gives a summary of dataset statistics.
The training set comes from a Chinese benchmark dataset named DetectRL-ZH~\cite{wu2024detectrl}, which is designed for detecting Chinese text produced by LLMs. The test set contains entirely out-of-domain data in terms of both content domain and LLM source. The evaluation covers three scenarios: normal, attack, and varying lengths.
More details refer to the official documentation provided by the task organizers~\cite{nlpcc2025task1}.

\vspace{-10pt}
\begin{table}
\caption{Details of the dataset provided by NLPCC2025 Shared Task 1.}
\label{dataset}
\centering
\begin{tabular}{l|r|r|r|c|c|c}
\hline
\multirow{2}{*}{Split} & \multicolumn{3}{c|}{Label Counts} & \multicolumn{3}{c}{Data Type} \\
\cline{2-7}
 & LLM & Human & Total & Normal & Attack & Varying Length \\
\hline
train & 24300 & 8100 & 32400 & $\checkmark$ & $\times$ & $\times$ \\
dev & 1700 & 1100 & 2800 & $\checkmark$ & $\times$ & $\times$ \\
test & 5500 & 5500 & 11000 & $\checkmark$ & $\checkmark$ & $\checkmark$ \\
\hline
\end{tabular}
\end{table}
\vspace{-20pt}

\subsubsection{System Implementation Details.} We adopt the same base detection models as described in Section 3.1. Specifically, we employ four \textbf{rule-based methods}: SpecialToken, CommonPhrase, SentenceSegment, and ConsecutivePunctuation.
For \textbf{training-free methods}, we used Binoculars~\cite{hans2024spotting}, Fast-DetectGPT~\cite{bao2024fastdetectgpt}, and CommonToken. Binoculars is implemented with the Qwen2.5-7B~\cite{qwen2.5} series as its backbone, while Fast-DetectGPT is tested with both the Qwen2.5-7B and GLM-4-9B~\cite{glm2024chatglm} series. The tokenizer for CommonToken is based on GLM-4-9B. We also calculate classification thresholds for Binoculars and Fast-DetectGPT under different text lengths using the training set.
For \textbf{training-based methods}, we fine-tune Chinese BERT and RoBERTa~\cite{cui-etal-2020-revisiting} models tailored to varying text lengths, as well as their corresponding hybrid-feature variants. In addition, we employe LoRA~\cite{hu2022lora} fine-tuning via LlamaFactory~\cite{zheng-etal-2024-llamafactory}, applying instruction tuning to Qwen2.5-7B-Instruct and GLM-4-9B-Chat, with Qwen2.5-7B-Instruct further adapted into several length-specific variants. The models trained on texts containing 110 to 150 words are referred to as the “short text” model, while the models trained on texts containing 55 to 75 words are referred to as the “extreme short text” model.
For the \textbf{decision support module}, we employ Qwen2.5-72B-Instruct to identify back-translated samples via ICL. The few-shot demonstration examples are constructed by applying Chinese $\rightarrow$ English $\rightarrow$ Chinese back-translation to five samples from the training set, using the mbart-large-50-many-to-many-mmt model~\cite{tang2020multilingual}.
The hyperparameters used in our system can be found in Table~\ref{hyperparam}.
Due to space limitations, more details can be found in the code repository.

\subsubsection{Metrics.} According to the official setup of NLPCC2025 Shared Task 1~\cite{nlpcc2025task1}, we adopt macro F1-score as our primary evaluation metric. The reliability of each model is reflected by its macro F1 score on the entire test set, which includes a variety of adversarial samples to simulate real-world scenarios.

\begin{table}
\caption{The hyper-parameters used in our system. The order of the weights $w$ corresponds to the order of the base models listed in Table~\ref{main_result}.}
\label{hyperparam}
\centering
\begin{tabular}{l|c|c|c|c|c|c|c|c|c|c|c|c|c|c|c|c|c|c|c|c}
\hline
text type & $w_1$ & $w_2$ & $w_3$ & $w_4$ & $w_5$ & $w_6$ & $w_7$ & $w_8$ & $w_9$ & $w_{10}$ & $w_{11}$ & $w_{12}$ & $w_{13}$ & $w_{14}$ & $w_{15}$ & $w_{16}$ & $w_{17}$ & $w_{18}$ & $\lambda$ & $\tau$ \\
\hline
ext. short & 0 & 10 & 0 & 10 & 10 & 60 & 60 & 55 & 60 & 0 & 0 & 0 & 0 & 0 & 95 & 400 & 10 & 0 & 250 & 0\\
short & 0 & 10 & 0 & 10 & 40 & 40 & 40 & 35 & 40 & 0 & 0 & 0 & 0 & 40 & 95 & 400 & 40 & 0 & 150 & 0\\
medium & 0 & 10 & 0 & 10 & 40 & 40 & 40 & 35 & 40 & 0 & 0 & 0 & 0 & 100 & 80 & 90 & 40 & 0 & 0 & 0\\
general & 0 & 10 & 10 & 10 & 40 & 70 & 70 & 70 & 75 & 50 & 60 & 0 & 85 & 400 & 40 & 60 &  80 & 0 & 0 & 0\\
\hline
\end{tabular}
\end{table}

\subsection{Evaluation Results}
We evaluate the LLM-generated text detection system implemented based on the EnsemJudge framework on the official test set. Table~\ref{main_result} reports the macro F1 scores of the overall system as well as each base model across different test set partitions. Specifically, “Mixed”, “Paraphrase” and “Perturbation” denote three types of adversarial attacks, while “64”, “128”, “256” and “512” indicate four different text lengths. Labels such as “5001-6000” represent the ID ranges of the corresponding subsets within the test set.

\vspace{-10pt}
\begin{table}
\caption{Main results on test set.}
\label{main_result}
\centering
\resizebox{\textwidth}{!}{
\begin{tabular}{l|c|c|c|c|c|c|c|c|c}
\hline
\multirow{3}{*}{Model Name} & \multirow{2}{*}{All} & \multirow{2}{*}{Normal} & \multicolumn{3}{c|}{Attack} & \multicolumn{4}{c}{Varying Length (Short Text)} \\
\cline{4-10}
 & & & Mixed & Paraphrase & Perturbation & 64 & 128 & 256 & 512 \\
\cline{2-10}
 & 1-11000 & 1-4000 & 4001-5000 & 5001-6000 & 6001-7000 & 7001-8000 & 8001-9000 & 9001-10000 & 10001-11000 \\
\hline
SpecialToken & 0.3334 & 0.3333 & 0.3338 & 0.3333 & 0.3333 & 0.3333 & 0.3333 & 0.3333 & 0.3333 \\
ConsecutivePunctuation & 0.3377 & 0.3367 & 0.3395 & 0.3400 & 0.3400 & 0.3333 & 0.3356 & 0.3400 & 0.3400 \\
CommonPhrase & 0.5299 & 0.5926 & 0.5122 & 0.6306 & 0.5325 & 0.3813 & 0.4020 & 0.4324 & 0.4681 \\
SentenceSegment & 0.5899 & 0.5919 & 0.6225 & 0.6361 & 0.6387 & 0.4864 & 0.5373 & 0.5627 & 0.6243 \\
CommonToken(GLM-4-9B) & 0.7327 & 0.7173 & 0.5958 & 0.8757 & 0.7584 & 0.6876 & 0.7065 & 0.7612 & 0.7744 \\
Fast-DetectGPT(Qwen2.5-7B/Qwen2.5-7B-Instruct) & 0.8298 & 0.9387 & 0.6400 & 0.7139 & 0.4355 & 0.8289 & 0.8419 & 0.8748 & 0.9230 \\
Fast-DetectGPT-analytical(Qwen2.5-7B-Instruct) & 0.8146 & 0.9775 & 0.4442 & 0.3808 & 0.3367 & 0.8036 & 0.8680 & 0.9160 & 0.9600 \\
Fast-DetectGPT-analytical(GLM-4-9B-Chat) & 0.8140 & 0.9414 & 0.6948 & 0.5924 & 0.5485 & 0.7212 & 0.8120 & 0.8455 & 0.9079 \\
Binoculars(Qwen2.5-7B-Instruct/Qwen2.5-7B) & 0.8317 & 0.9890 & 0.4828 & 0.3777 & 0.3444 & 0.8360 & 0.8880 & 0.9350 & 0.9800 \\
ChineseBERT & 0.7719 & 0.8592 & 0.5567 & 0.8561 & 0.8197 & 0.4624 & 0.5687 & 0.7636 & 0.8713 \\
ChineseRoBERTa & 0.7701 & 0.8426 & 0.5141 & 0.7041 & 0.8476 & 0.5566 & 0.6858 & 0.7773 & 0.9177 \\
ChineseRoBERTa(Extreme Short Text) & 0.8377 & 0.8694 & 0.5091 & 0.8583 & 0.8645 & 0.7751 & 0.8286 & 0.9057 & 0.9250 \\
GLM-4-9B-Chat-LoRA & 0.8907 & \underline{0.9895} & 0.7984 & 0.9529 & \underline{0.9920} & 0.3400 & 0.6127 & 0.9428 & \underline{0.9970} \\
Qwen2.5-7B-Instruct-LoRA & 0.9409 & \textbf{1.0000} & \textbf{0.9970} & \underline{0.9589} & \textbf{0.9990} & 0.4544 & 0.8444 & \underline{0.9890} & \textbf{1.0000} \\
Qwen2.5-7B-Instruct-LoRA(Extreme Short Text) & 0.9057 & 0.9431 & 0.6562 & 0.7641 & 0.8827 & 0.9227 & 0.9479 & 0.9850 & 0.9910 \\
Qwen2.5-7B-Instruct-LoRA(Short Text) & \underline{0.9572} & 0.9605 & \underline{0.9064} & 0.9359 & 0.9610 & \underline{0.9470} & \underline{0.9700} & 0.9790 & 0.9870 \\
HybridFeatureRoBERTa & 0.8555 & 0.9214 & 0.5434 & 0.8545 & 0.9020 & 0.7232 & 0.8269 & 0.8901 & 0.9309 \\
HybridFeatureRoBERTa(Extreme Short Text) & 0.8204 & 0.8401 & 0.4673 & 0.7974 & 0.8274 & 0.8081 & 0.8510 & 0.9160 & 0.9148 \\
\textbf{EnsemJudge} & \textbf{0.9922} & \textbf{1.0000} & \textbf{0.9970} & \textbf{0.9870} & \textbf{0.9990} & \textbf{0.9590} & \textbf{0.9780} & \textbf{0.9940} & \textbf{1.0000} \\
\hline
\end{tabular}
}
\end{table}
\vspace{-10pt}

\begin{figure}[htbp]
\centering
\begin{adjustbox}{width=\textwidth}
\begin{tikzpicture}
\begin{axis}[
    colormap/jet,
    colorbar,
    point meta min=0.0,
    point meta max=1.0,
    xlabel={Subset},
    ylabel={Model},
    xtick={0,...,8},
    xticklabels={All,Normal,Mixed,Paraphr.,Perturb.,Len-64,Len-128,Len-256,Len-512},
    ytick={0,...,18},
    yticklabels={
        SpecialToken,
        ConsecutivePunctuation,
        CommonPhrase,
        SentenceSegment,
        CommonToken(GLM-4-9B),
        Fast-DetectGPT(Qwen2.5-7B/Qwen2.5-7B-Instruct),
        Fast-DetectGPT-analytical(Qwen2.5-7B-Instruct),
        Fast-DetectGPT-analytical(GLM-4-9B-Chat),
        Binoculars(Qwen2.5-7B-Instruct/Qwen2.5-7B),
        ChineseBERT,
        ChineseRoBERTa,
        ChineseRoBERTa(Extreme Short Text),
        GLM-4-9B-Chat-LoRA,
        Qwen2.5-7B-Instruct-LoRA,
        Qwen2.5-7B-Instruct-LoRA(Extreme Short Text),
        Qwen2.5-7B-Instruct-LoRA(Short Text),
        HybridFeatureRoBERTa,
        HybridFeatureRoBERTa(Extreme Short Text),
        EnsemJudge
    },
    x tick label style={rotate=270, anchor=west},
    enlargelimits=false,
    tick label style={font=\small},
]
\addplot [matrix plot*, point meta=explicit, mesh/rows=19,mesh/cols=9] table [meta=val] {
x y val
0 0 0.3334
1 0 0.3333
2 0 0.3338
3 0 0.3333
4 0 0.3333
5 0 0.3333
6 0 0.3333
7 0 0.3333
8 0 0.3333
0 1 0.3377
1 1 0.3367 
2 1 0.3395 
3 1 0.34 
4 1 0.34 
5 1 0.3333 
6 1 0.3356 
7 1 0.34 
8 1 0.34
0 2 0.5299 
1 2 0.5926 
2 2 0.5122 
3 2 0.6306 
4 2 0.5325 
5 2 0.3813 
6 2 0.402 
7 2 0.4324 
8 2 0.4681
0 3 0.5899 
1 3 0.5919 
2 3 0.6225 
3 3 0.6361 
4 3 0.6387 
5 3 0.4864 
6 3 0.5373 
7 3 0.5627 
8 3 0.6243
0 4 0.7327 
1 4 0.7173 
2 4 0.5958 
3 4 0.8757 
4 4 0.7584 
5 4 0.6876 
6 4 0.7065 
7 4 0.7612 
8 4 0.7744
0 5 0.8298 
1 5 0.9387 
2 5 0.64 
3 5 0.7139 
4 5 0.4355 
5 5 0.8289 
6 5 0.8419 
7 5 0.8748 
8 5 0.923
0 6 0.8146 
1 6 0.9775 
2 6 0.4442 
3 6 0.3808 
4 6 0.3367 
5 6 0.8036 
6 6 0.868 
7 6 0.916 
8 6 0.96
0 7 0.814 
1 7 0.9414 
2 7 0.6948 
3 7 0.5924 
4 7 0.5485 
5 7 0.7212 
6 7 0.812 
7 7 0.8455 
8 7 0.9079
0 8 0.8317 
1 8 0.989 
2 8 0.4828 
3 8 0.3777 
4 8 0.3444 
5 8 0.836 
6 8 0.888 
7 8 0.935 
8 8 0.98
0 9 0.7719 
1 9 0.8592 
2 9 0.5567 
3 9 0.8561 
4 9 0.8197 
5 9 0.4624 
6 9 0.5687 
7 9 0.7636 
8 9 0.8713
0 10 0.7701 
1 10 0.8426 
2 10 0.5141 
3 10 0.7041 
4 10 0.8476 
5 10 0.5566 
6 10 0.6858 
7 10 0.7773 
8 10 0.9177
0 11 0.8377 
1 11 0.8694 
2 11 0.5091 
3 11 0.8583 
4 11 0.8645 
5 11 0.7751 
6 11 0.8286 
7 11 0.9057 
8 11 0.925
0 12 0.8907 
1 12 0.9895 
2 12 0.7984 
3 12 0.9529 
4 12 0.992 
5 12 0.34 
6 12 0.6127 
7 12 0.9428 
8 12 0.997
0 13 0.9409 
1 13 1 
2 13 0.997 
3 13 0.9589 
4 13 0.999 
5 13 0.4544 
6 13 0.8444 
7 13 0.989 
8 13 1
0 14 0.9057 
1 14 0.9431 
2 14 0.6562 
3 14 0.7641 
4 14 0.8827 
5 14 0.9227 
6 14 0.9479 
7 14 0.985 
8 14 0.991
0 15 0.9572 
1 15 0.9605 
2 15 0.9064 
3 15 0.9359 
4 15 0.961 
5 15 0.947 
6 15 0.97 
7 15 0.979 
8 15 0.987
0 16 0.8555 
1 16 0.9214 
2 16 0.5434 
3 16 0.8545 
4 16 0.902 
5 16 0.7232 
6 16 0.8269 
7 16 0.8901 
8 16 0.9309
0 17 0.8204 
1 17 0.8401 
2 17 0.4673 
3 17 0.7974 
4 17 0.8274 
5 17 0.8081 
6 17 0.851 
7 17 0.916 
8 17 0.9148
0 18 0.9922 
1 18 1 
2 18 0.997 
3 18 0.987 
4 18 0.999 
5 18 0.959 
6 18 0.978 
7 18 0.994 
8 18 1
};
\end{axis}
\end{tikzpicture}
\end{adjustbox}
\caption{Model performance (macro F1-score) for our system and different base models.}
\label{fig:heat}
\end{figure}
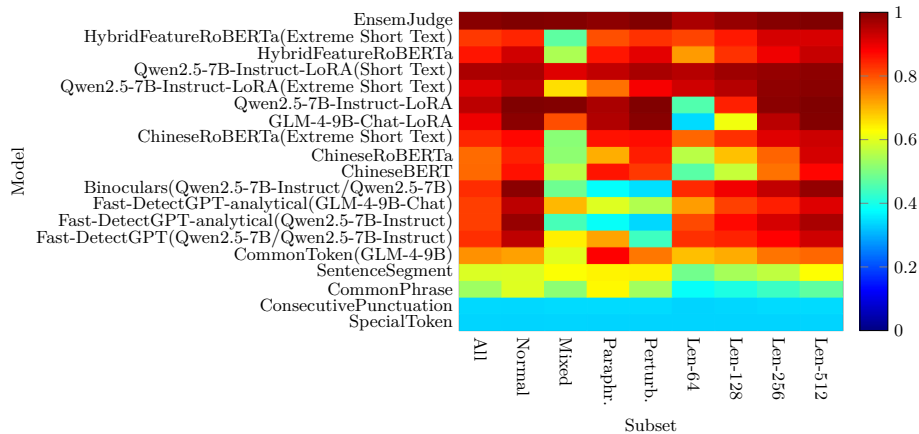

\noindent\textbf{Overall Performance.} Our system achieves the best performance on the overall test set as well as across all individual subsets. On non-adversarial samples, it reaches a perfect macro F1 score of 1.000, indicating completely correct predictions. Even on the full test set, the system achieves a macro F1 of 0.9922, with improvements up to 9.0\% on certain individual subsets, demonstrating strong reliability.
As illustrated in figure~\ref{fig:heat}, different categories of detection methods exhibited substantial performance differences. Rule-based methods are largely ineffective, with macro F1 scores not exceeding 0.6. Training-free methods perform moderately better, with scores around 0.8. Training-based methods show the best results, reaching up to 0.95. Furthermore, for methods of the same type, we observe similar patterns when confronted with various adversarial attacks. Their differences in performance primarily stem from specific implementation details.

\vspace{-30pt}

\subsubsection{Impact of Adversarial Attacks.}
We observe that different adversarial strategies have varying impacts on the models. Overall, the most effective adversarial method is mixing human-written and LLM-generated texts, while the most robust approach is fine-tuning LLMs. Notably, the fine-tuned Qwen2.5 model is barely affected by mixing or perturbation attacks with performance degradation under 1\%, and shows less than a 5\% drop under paraphrasing attacks. We speculate that this robustness may stem from Qwen2.5's superior pretraining mechanism, which grants it heightened sensitivity to global textual coherence and deep semantic structure, making it resistant to localized "camouflage." In contrast, these three types of attacks severely cripple training-free methods—e.g., Binoculars suffered over a 50\% drop in performance. Interestingly, our proposed CommonToken outperforms all other non-training-based approaches under paraphrasing and perturbation attacks, achieving gains of 16\% and 12\% respectively, and ranking just behind fine-tuned LLMs.

\subsubsection{Impact of Text Length.} We find that text length has a significant impact on the performance of all detection methods. As the length increases, the F1 score generally improves. However, this performance gain became marginal when the length exceeds approximately 500 characters. For example, in the best-performing base model—the fine-tuned Qwen2.5—the performance improves by 39.0\%, 14.5\%, and 1.1\% with increasing length.
Our experiments show that building specialized models for short texts can effectively mitigate the negative effects of limited input length. For instance, setting length-aware decision thresholds for training-free methods or training models specifically on short-text data for training-based methods are both beneficial strategies. On extreme-short text subsets, the short-text variants of RoBERTa, Qwen2.5, and our hybrid feature model outperform their regular counterparts by 21.9\%, 46.8\%, and 8.5\%, respectively.

\subsubsection{Impact of Model Implementations.}
We observe that even within the same category of detection methods, different implementation choices can lead to significant performance differences. For instance, the LoRA fine-tuned Qwen2.5-7B-Instruct consistently outperforms GLM-4-9B-Chat across all types of input texts, achieving an overall performance improvement of 5\%.
As for Fast-DetectGPT, using GLM-4-9B-Chat as the backbone leads to better results on adversarial samples, whereas Qwen2.5-7B-Instruct performs better on regular and variable-length texts. However, the overall performance gap between the two backbone choices is less than 0.1\%.

\section{Conclusion}
In this paper, we propose EnsemJudge, a reliable and extensible ensemble framework for LLM-generated text detection in real-world scenarios. By dynamically assigning voting strategies based on lightweight textual features and integrating a decision support module, EnsemJudge effectively leverages the strengths of diverse detection methods while mitigating their weaknesses. Our system is rigorously evaluated on the NLPCC2025 Shared Task 1 dataset, which includes both in-distribution and challenging out-of-distribution adversarial samples. The experimental results demonstrate that EnsemJudge achieves state-of-the-art performance, showing strong generalization and robustness across various text types and attack settings. This work highlights the importance of adaptive ensemble approaches for reliable LLM-generated text detection, especially in low-resource languages such as Chinese, and provides a practical foundation for future research and deployment in real-world applications.

\begin{credits}
\subsubsection{\ackname} This work is supported by the National Natural Science Foundation of China (No.U2336202).
\end{credits}

\bibliographystyle{splncs04_unsort}
\bibliography{refs}

\end{document}